\documentclass[compsoc]{IEEEtran}

\usepackage[utf8]{inputenc}

\usepackage[pdftex]{graphicx}
\usepackage{amsmath,amssymb}
\usepackage{url}
\usepackage{path}
\usepackage{xcolor}
\usepackage[nocompress]{cite}
\usepackage{booktabs, multicol, multirow}
\usepackage[caption=false]{subfig}
\usepackage{longtable}
\usepackage[all]{xy}
\usepackage{amssymb,amsmath}
\usepackage{latexsym}
\usepackage[ruled,vlined,linesnumbered]{algorithm2e}

\definecolor{newcolor}{rgb}{.8,.349,.1}

\begin{document}

\title{Bipartite Graph Matching for Keyframe Summary Evaluation}

\author{Iain~A.~D.~Gunn, Ludmila~I.~Kuncheva, and Paria Yousefi,\IEEEcompsocitemizethanks{
\IEEEcompsocthanksitem I.~Gunn is with the Department of Computer Science, Middlesex University, The Burroughs, London NW4 4BT (email: i.gunn@mdx.ac.uk).
\IEEEcompsocthanksitem L.~Kuncheva and P.~Yousefi are with the School of Computer Science, Bangor University, Dean Street, Bangor, Gwynedd, LL57 2NJ, UK (email: \{l.i.kuncheva,p.yousefi\}@bangor.ac.uk).
}
\thanks{}
}

\IEEEtitleabstractindextext{%
\begin{abstract}
A keyframe summary, or ``static storyboard", is a collection of frames from a video designed to summarise its semantic content. Many algorithms have been proposed to extract such summaries automatically.  How best to evaluate these outputs is an important but little-discussed question.  We review the current methods for matching frames between two summaries in the formalism of graph theory. Our analysis revealed different behaviours of these methods, which we illustrate with a number of case studies. Based on the results, we recommend a greedy matching algorithm due to Kannappan et al.
\end{abstract}

}
\maketitle

\IEEEdisplaynontitleabstractindextext

\IEEEraisesectionheading
{\section{Introduction}
\label{introduction}}

Keyframe selection is the selection of a set of representative fames from a video to provide a summary of its semantic content for human interpretation \cite{Truong2007}. Many algorithms have been proposed to extract keyframe summaries automatically, and the question of the evaluation of the outputs of these algorithms has naturally arisen.  Output summaries are often evaluated by a user ``taste-test'', where a survey is conducted in which participants are asked to rank or score the outputs of several algorithms.  In contrast to these methods, there is potentially great value in establishing automatic evaluation schemes based on ``ground truths''.  Given user-created keyframe summaries (the ``ground truths'') associated to some standard data sets, the outputs of proposed new algorithms can be compared to these. Measures of performance can then be defined in terms of how well the algorithm output matches the ground truths.  This has the advantage that a new algorithm can easily be compared to any existing algorithm simply by comparing the performance scores with respect to the ground truth. In taste-test methods, on the other hand, any algorithm not included in the original experiment cannot subsequently be compared to the proposed algorithm except by assembling a whole new user survey.  A method based on published ground truths makes it possible to perform experimental studies without the time and expense involved in conducting a user survey, enabling larger experiments with greater repeatability.

The utility of evaluation methods based on ground-truths rests on having a suitable measure of the performance of each algorithm compared with the ground truth.  A measure of distance between the output keyframe set and the ground-truth keyframe set is therefore sought.  It is natural to think of using existing concepts of distance between sets such as the Hausdorff distance, but these may unduly penalise a small number of distant elements in one set (see our discussion below).  For, perhaps, this reason, de Avila et al. \cite{DeAvila2011} proposed a scheme based on a count of pairings between frames, which has gained some popularity.  The subjects of the present study are this scheme, similar schemes based on frame pairings, and potential novel schemes like them.

The goals of the present work are as follows.  First, to present clearly the method of de Avila et al.\ and similar competing methods in the language of graph theory. Graph theory provides a natural formalism for matching problems, in which matching problems similar to the one at hand have previously been theoretically studied.  Second, to discuss the different behaviours of these methods, 
and to illustrate the contrasting behaviour with examples.  Third, to consider which of these methods might be best suited to the task of evaluating keyframe summaries.

In section \ref{sec:StandardTheory}, we review some standard theoretical ideas which have been applied by some authors to the problem of comparing keyframe sets.  In section \ref{sec:thresh}, we review the thresholded matching scheme proposed by de Avila et al., and certain alternative matching algorithms which have been proposed by other authors which are suitable for use within that framework. In section \ref{sec:discussion} we discuss the different behaviours of these alternative matching algorithms, with examples.  In section \ref{sec:directions} we briefly indicate some possible future directions for development of keyframe matching methods, and in section \ref{sec:conclusion} we present our conclusions.

\section{Similarity measures using standard approaches}
\label{sec:StandardTheory}

For completeness and context, we briefly review some standard set-theoretic concepts of distance between sets which have been used for comparing keyframe sets.  We then introduce the graph-theoretic formalism in which we will present and discuss the alternative evaluation algorithms which are the main subject of this work.

Note that all these concepts of distance between sets of frames depend on an underlying concept of distance between individual frames.  The choice of inter-frame distance measure is a significant question itself, and beyond the scope of the present work.  Throughout this work we will use the underlying distance used by de Avila et al., which is based on a 16-bin histogram of the hue values of the pixels of each frame~\cite{DeAvila2011}.

\subsection{Set-theoretic approaches}

Given the problem of evaluating ``closeness'' between a candidate key-frame set and a provided ground-truth set of frames, it is natural to think of using the pre-existing concepts of Hausdorff distance or Hausdorff semi-distance between sets. The Hausdorff distance between two non-empty sets $X$, $Y$ of elements of a metric space with metric $d$ is defined as
\begin{equation}
 d_H(X,Y) = \mathrm{max}( \underset{x \in X}{\mathrm{sup}} \enskip \underset{y \in Y}{\mathrm{inf}} \enskip d(x,y), \enskip \underset{y \in Y}{\mathrm{sup}} \enskip \underset{x \in X}{\mathrm{inf}} \enskip d(x,y)).
\end{equation}
That is, informally, the Hausdorff distance is the worst case of best cases: for every point in the two sets, find the distance to the nearest point in the other set; the greatest of these distances is the Hausdorff distance between the two sets.  It is determined totally by this worst-case element; all information about closer elements is lost.

The Hausdorff semi-distance is defined as
\begin{equation}
 d_S(X,Y) = \underset{x \in X}{\mathrm{sup}} \enskip \underset{y \in Y}{\mathrm{inf}} \enskip d(x,y).
\end{equation}
Note the asymmetry with respect to the order of the two sets $X$,$Y$, which means the Hausdorff semi-distance is not a metric.  This does not in itself prevent the Hausdorff semi-distance from being a suitable score for the present purpose of evaluation: the different status of the candidate keyframe set and the ground truth to which it is being compared could make an asymmetric measure appropriate.

Several authors, following the ``Fidelity'' idea of Chang et al \cite{Chang1999}, have used these measures to compute a distance from the keyframe set to the full video sequence which it summarises. The other use is to compute a distance to a ground-truth keyframe set from a candidate set.  However, as mentioned, the Hausdorff (semi-)distance may not be the most appropriate measure for distance between keyframe sets, as it gives undue weight to the worst-case distance among elements of the set.  In evaluating a keyframe summary, it is natural to wish to give weight to the performance with respect to better-performing candidate keyframes.  Indeed, it is the very distant members of the two sets whose distance should be discounted: if a candidate keyframe has ``missed'', being close to no ground-truth keyframe, it should not matter how much it has missed by in whatever metric is being used in the feature space.  If we compare a ground-truth keyframe showing, say, a tree, first to a candidate keyframe showing a cat, then to a candidate keyframe showing a dog, it would be entirely pointless to use the metric of the feature space to attempt to answer the question of which of these candidate keyframes is \emph{more} wrong as a match for the tree.  The distance information is spurious in so far as it purports to discriminate between keyframes which are both definitely inappropriate.  Therefore it is desirable that a concept of distance between keyframe sets should have a provision for simply discounting such ``missed'' keyframes.  This explains the thresholding procedure described in section \ref{sec:thresh} below. 

A possible improvement on the Hausdorff distance is to use a weighted average of the best distance from each point in the first set to a point in the second set: this is the measure used by, for example, Li and Merialdo \cite{Li2010}.  This approach will still be strongly affected by outliers, though not to the extreme extent that the Hausdorff (semi-)distance is.

In addition to giving strong undue weight to the most distant elements, the Hausdorff semi-distance and weighted average distance share a second serious failing: they do not penalise the case in which many points of one set are close to one point of the second set; that is, they do not detect whether the coverage of one set by the other is good.  The methods which have gained popularity, and which are discussed in the remainder of this paper, address this problem by pairing off the elements of the two sets, through one of various matching processes.

\subsection{Graph-theoretic approach}

The methods for comparing keyframe sets which have gained most popularity, especially the method of de Avila et al., are based on the idea that each keyframe in a keyframe set under evaluation should be paired with at most one keyframe in the ground-truth set and vice versa.  This avoids the pitfall discussed above, in which two sets might falsely appear to be good matches if many elements of one set are close to one or a few elements of the other.  It is clear from the nature of a keyframe summary that multiple frames from one summary should not be considered as equivalent to a single frame of another summary: in such a case, either the first summary is suffering from redundancy or the second summary has failed to summarise a recurring concept captured by the first.

Given that each frame in a set ought to correspond to at most one frame in the set to which it is being compared, we have a formal problem of exclusively pairing elements of one set to elements of a second set.  This is, in the language of graph theory, a bipartite graph matching problem.


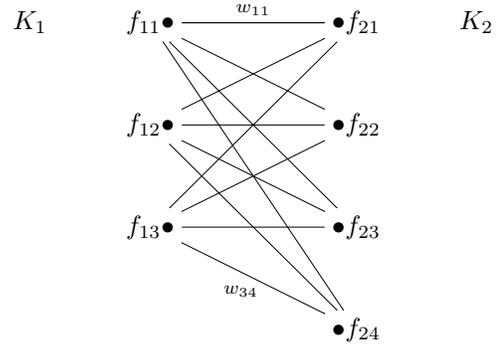
\begin{figure}

\[
 \xymatrix{K_1&f_{11}\bullet
\ar@{-}[rr]^{w_{11}} \ar@{-}[drr] \ar@{-}[rr] \ar@{-}[ddrr] \ar@{-}[dddrr] 
&& \bullet f_{21} & K_2\\
&f_{12}\bullet \ar@{-}[urr] \ar@{-}[drr] \ar@{-}[rr] \ar@{-}[ddrr]  
&&\bullet f_{22}&\\
&f_{13}\bullet
\ar@{-}[uurr] \ar@{-}[urr] \ar@{-}[rr] \ar@{-}[drr]_{w_{34}} &&\bullet f_{23}&\\
&&&\bullet f_{24}&\\}
\]
	\caption{A graph-theoretic framework for comparing keyframe sets.  Each frame in $K_1$ corresponds to a vertex in the left part of the graph, and each frame in $K_2$ corresponds to a vertex in the right part of the graph.  Each vertex is connected to all vertices of the opposite keyframe set and to no vertices of the same keyframe set. The weight of each edge is the distance between the frames in a feature space. }
	\label{fig:BipartiteGraphDefinition}
\end{figure}

Let each keyframe in each set correspond to a vertex of the graph.  The form of the graph is illustrated in figure \ref{fig:BipartiteGraphDefinition}.  From each vertex, let there be edges connecting to all vertices corresponding to keyframes of the other set (this gives a complete bipartite graph).  Let the weight of each edge be the distance between the two keyframes corresponding to the vertices it joins (where the distance between keyframes is defined using a previously chosen feature space and metric, in our case the choices of de Avila et al.).

A \textit{matching}, in graph theory, is a set of edges with no common vertices.  Thus, a matching for a bipartite graph as defined above gives a set of pairings from the first keyframe set $K_1$ to the second keyframe set $K_2$, such that each keyframe is paired to at most one element of the other set. The task now is to find a suitable matching for the graph thus defined, which can then be used to form an appropriate measure of distance between the sets.

One obvious approach, if the sets are of equal cardinality, is to consider complete matchings (in which every vertex is assigned a pair) and determine the complete matching of lowest total weight.  This method has the advantage of giving a distance between sets expressed in the metric of the space in which the frames reside, as does the Hausdorff (semi-)distance.  This matching problem is solved by the Hungarian algorithm, a standard method from graph theory.  This method is used by, for example, Khosla et al. \cite{Khosla2013}.


This approach solves one problem of the (semi-)Hausdorff and average-distance approaches: it does not allow a set which clusters closely around one or a few elements of another set to count as being a good match.  However, it still has the problem of excessively penalising distant outliers.  This second problem is additionally addressed by the thresholded-matching method described below.  

\section{Thresholded matching}
\label{sec:thresh}

The algorithms we consider in this section match a keyframe in one summary to at most one keyframe in the other summary, and so can usefully be described in the graph-theoretic formalism as matching algorithms.  But they do not find complete matchings, unlike the Hungarian algorithm.

Following de Avila et al. \cite{DeAvila2011}, many authors (e.g. \cite{Ejaz2012}, \cite{Gong2014}, \cite{Mei2015}) have taken an approach in which keyframes from the summary under evaluation are paired to sufficiently close keyframes in the target ground-truth, and the number of such pairs is taken as the basis for measures of the performance of the algorithm (e.g. recall, precision, and F-measure, which are simple scalings of the number of pairs by the sizes of the keyframe sets).  In contrast to the methods discussed so far, these measures do not give a distance in terms of the metric of the underlying feature space.

In the scheme of de Avila et al., a threshold $\theta$ is specified for the distance between frames (in the underlying feature space in which the frames are represented).  Two frames $f_a$ and $f_b$, represented respectively as $a$ and $b$ in the space of interest $\mathbb{R}^n$ are considered sufficiently similar to be paired if the distance between $a$ and $b$ is smaller than this pre-set threshold: $d(a,b)<\theta$. A matching is then sought between the elements of the candidate keyframe set and the elements of the ground-truth keyframe set, considering only pairings involving distances below the specified threshold.  That is, a matching is found for a graph containing only those edges whose weights are under the threshold.  Various measures of the closeness of the sets are then calculated based on the cardinality of this matching.

It is crucially important to note that the number of pairs one finds will in general depend on the algorithm one uses to perform the matching!  Alarmingly, many authors do not comment on how the matching is performed and appear not to have reflected that different algorithms will in general give different solutions.

The question of which algorithm is most desirable for the task at hand is more subtle than it may at first appear.  One's first thought might be that having performed the thresholding described above, we have replaced the weighted matching problem with an unweighted matching problem: that is, that the question is now ``how many pairings can I make, given that I've thrown out the unsuitable pairings?''. This unweighted problem would then be optimally solved by any algorithm returning a maximal matching, such as the Hopcroft-Karp algorithm.  However, it may be beneficial to make further use of the weight information, beyond the removal of the clearly unacceptable over-threshold pairings.  A maximal matching algorithm, which ignores the relative weights among sub-threshold edges, will always prefer more worse pairings to fewer better ones, which may be undesirable even given that over-threshold pairings have been removed: if two frames are supposed to be paired, but a better pairing exists for at least one of them, the sub-optimal pairing may well be spurious.  This is because typically, elements of a key-frame summary will cover distinct semantic content: an ideal keyframe summary is free of redundancy.  So if an algorithm proposes to pair a frame in one summary to a frame in another summary other than its best match in that summary, the chances may be low that this pairing is really correct in the sense of matching frames with similar semantic content, unless the second summary is a poor summary with much redundancy (particularly unlikely for user-generated summaries which are used as ground-truths).  The major exception would be the case of a video stream with a recurring view, which it might be appropriate to show multiple times in a summary.

So, thresholding followed by treating the problem as an unweighted matching might be a sub-optimal approach.  
A simple but effective way to use the distance information following thresholding is to use a greedy algorithm, Algorithm~\ref{GreedyMatching}.

\begin{algorithm}
	\DontPrintSemicolon
	
	\medskip
	\KwIn{The distance matrix $D$ between keyframe summaries $K_1$ and $K_2$, and threshold $\theta$.}
	
	\medskip
	\KwOut{Number of pairings $m$.}

	\medskip
	$m\gets 0.$
	Find the smallest distance $d_{\min} = \min D$. 
	\While {$d_{\min}< \theta$} 
	{Increment the number of matches, $m \gets m+1$.
	
	Remove the row and the column of the matched elements from $D$.
	
	Find the smallest distance from the remaining matrix $d_{\min} = \min D$.} 

	\caption{Greedy Matching}
	\label{GreedyMatching}
\end{algorithm}

The other algorithms which we shall describe in this section are alternative algorithms for performing the matching task, with the same or a similar scheme for thresholding based on distance in a feature space.  We shall be considering these algorithms within the framework of de Avila et al., thus not precisely as their authors intended, which could impact their performance.

Mahmoud \cite{Mahmoud2014} proposes a method which uses an even simpler algorithm than the greedy algorithm above to calculate the number of pairings (Algorithm~\ref{MahmoudMatching}).  For each keyframe in the candidate summary, he simply iterates over all frames of the ground-truth summary, and pairs the candidate keyframe to the first ground-truth keyframe which passes the thresholding criteria.  (He thresholds two distances corresponding to two sets of features, but the approach is otherwise broadly similar to the de Avila model.)

\begin{algorithm}
	\DontPrintSemicolon
	
	\medskip
	\KwIn{Keyframe summaries $K_1$ and $K_2$ arranged in temporal order, and threshold $\theta$.}
	
	\medskip
	\KwOut{Number of pairings $m$.}

	\medskip
	$m\gets 0.$

	\For {each frame  $a \in K_1$} 
	{\For {each frame in $b \in K_2$}
	{\If {this pair of frames is close enough to be paired, $d(a,b)<\theta$,}
		{Increment the number of pairings, $m \gets m+1$.
	
		Remove $a$ from $K_1$ and $b$ from $K_2$.

		Break.}
	}
	}

	\caption{Algorithm of Mahmoud \cite{Mahmoud2014}}
	\label{MahmoudMatching}
\end{algorithm}


An interesting alternative approach to the matching problem is put forward by Kannappan et al.~\cite{Kannappan16}. In their approach, a keyframe from the candidate set and  a keyframe from the ground truth are matched only if they are each the other's best possible match: Algorithm~\ref{KannappanMatching}.  In their implementation, the set of matched pairs is subsequently thresholded using a different concept of pairwise frame distance from that used to form the matches.  We have modified this procedure to make it the equivalent of the de Avila et al.\ thresholding, by using the same distance metric for thresholding as for finding the pairings.

\begin{algorithm}
	\DontPrintSemicolon
	
	\medskip
	\KwIn{The distance matrix $D$ between keyframe summaries $K_1$ and $K_2$, and threshold $\theta$.}

	\medskip
	\KwOut{Number of pairings $m$.}

	\medskip
	Initialise a set of pairings $M\gets \emptyset.$

	\For {each frame  $a \in K_1$} 
	{\For {each frame $b \in K_2$}
	{\If {$b = \arg \min_{b'\in K_2} d(a,b')$ and
		$a = \arg \min_{a'\in K_1} d(a',b)$}
		{Add the pair to the matching set $M\gets M\cup\{(a,b)\}$.}
	}
	}
	Remove from $M$ all pairs for which $d(a,b) \geq \theta$.

   $m\gets |M|.$

	\caption{Algorithm of Kannappan et al.~\cite{Kannappan16}}
	\label{KannappanMatching}
\end{algorithm}

%

We now present a simple example of a small bipartite graph to illustrate that Algorithms 1--3, and an algorithm returning a complete matching, will in general return different matchings of different cardinalities, \emph{even when using the same distance metric and threshold criterion.}

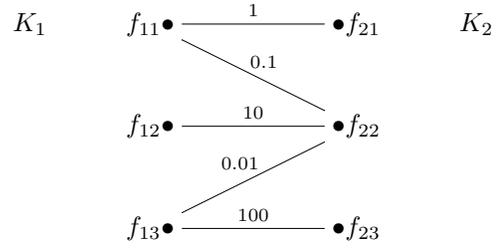
\begin{figure}

\[
 \xymatrix{K_1&f_{11}\bullet
\ar@{-}[rr]^{1} \ar@{-}[drr]^{0.1} && \bullet f_{21} & K_2\\
&f_{12}\bullet \ar@{-}[rr]^{10} &&\bullet f_{22}&\\
&f_{13}\bullet\ar@{-}[urr]^{0.01} \ar@{-}[rr]^{100} &&\bullet f_{23}&\\
}
\]

	\caption{Example of a small bipartite graph to illustrate behaviour of matching algorithms.  Each of the algorithms will return a different matching for this graph.  The numbers in the figure give the weights of the five edges.}
	\label{fig:BipartiteGraphExample}
\end{figure}

Consider the graph illustrated in Figure \ref{fig:BipartiteGraphExample}.  Suppose the weights of the edges, as given by the numbers in the figure, are all below the threshold.  Consider first the Greedy algorithm.  The greedy algorithm will find first the edge of lowest weight, 0.01, and will then add the only remaining possible pairing, the edge of weight 1.  This algorithm therefore finds a matching of cardinality 2. A complete matching, as returned by e.g. the Hopcroft-Karp algorithm~\cite{West}, is the matching of greatest cardinality: in this case, this is the matching of cardinality 3 given by the set of edges of weights 1, 10, and 100.  The output of the algorithm of Mahmoud is not stable; it depends on the order in which the vertices are stored in memory and presented to the for loop.  (This is not necessarily inappropriate given that keyframe collections do tend to come with an inherent temporal ordering, being derived from video.) Mahmoud's algorithm may find the matching of cardinality 2 found by the greedy algorithm; it may find the other matching of cardinality 2, given by the set of edges of weights 0.1 and 100; or it may find the matching of cardinality 3.  The algorithm of Kannappan et al. finds a matching of cardinality 1, consisting of the set containing only the edge of weight 0.01.

%
%
%
%

\section{Algorithm behaviour}
\label{sec:discussion}

Which algorithm should be used to form the matchings for keyframe evaluation?  The one which does the best job of pairing frames which truly share the same semantic content.  We present now some examples which illustrate that the conservative approach of the algorithm of Kannappan et al. may give a better matching than the competing algorithms.  The first example will also illustrate why simply imposing a more conservative (i.e. lower) threshold value will not always achieve as good a result as using a higher threshold with a more conservative matching algorithm.

These examples were made using the VSUMM project repository, which provides the outputs of five summarisation algorithms and five human-made ground-truth sets, for each of 50 videos.  The five summarisation algorithms considered are: DT (``Delauanay Triangulation'', \cite{Mundur2006}); OV (``Open Video'' project, summaries based on \cite{DeMenthon}); STIMO (``STIll and MOving Video Storyboard for the Web Scenario'' \cite{Furini2010}); and VSUMM1 \& VSUMM2 (``Video SUMMarization'' \cite{DeAvila2011}). Remember that we are considering these algorithms with the thresholding framework, feature space, and metric of de Avila et al., not as they were originally put forward.

\begin{figure*}
  \centering
  \subfloat[Greedy, $\theta=0.5$.]{
	\includegraphics[height=\linewidth]{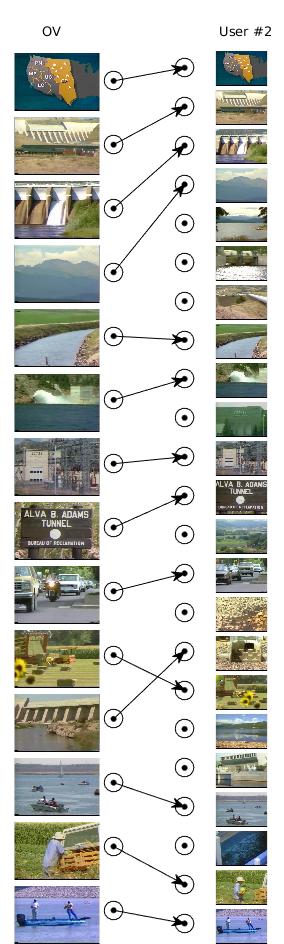}
	\label{fig:Vid8Greedy0.5}
  }
  \subfloat[Kannappan, $\theta = 0.5$.]{
	\includegraphics[height=\linewidth]{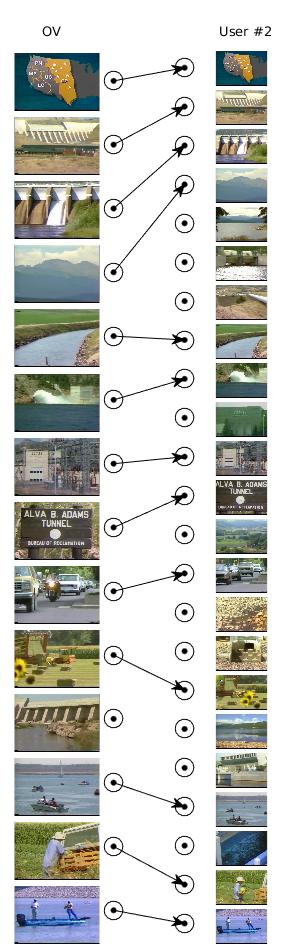}
	\label{fig:Vid8Kannappan0.5}
  }
  \subfloat[Greedy, $\theta = 0.3$]{
	\includegraphics[height=\linewidth]{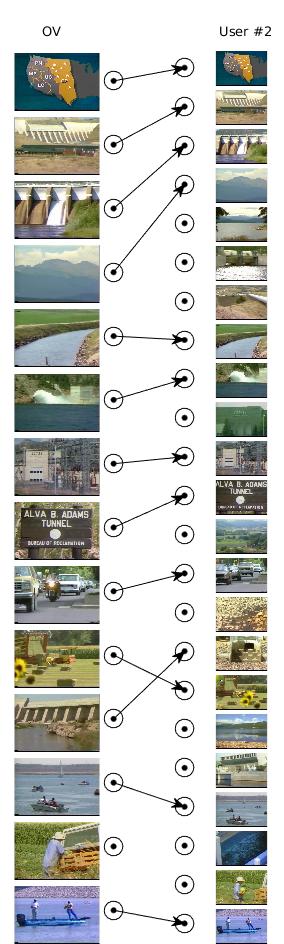}
	\label{fig:Vid8Greedy0.3}
  }
  \caption{Matchings found for Video 28}
  \label{fig:Vid8}
\end{figure*}

Figure \ref{fig:Vid8} shows matchings between the output of the OV algorithm and the summary provided by user number 2 for video number 28.  With the standard threshold value of 0.5, the algorithm of Kannappan et al. finds 13 true pairings between elements of the two summaries (figure \ref{fig:Vid8Kannappan0.5}).  The Greedy algorithm of de Avila et al. finds these 13 true pairings, and also one spurious pairing (the arrow that crosses another going upwards in figure \ref{fig:Vid8Greedy0.5}).  Thus, the conservative approach of the Kannappan et al. algorithm has led to a better matching with this threshold.  Note that attempting to enforce a conservative matching by using the greedy algorithm with a lower threshold leads to poorer results: if the threshold is reduced to 0.3, both algorithms lose one correct pairing (the second-to-bottom frames), but the Greedy algorithm still finds the spurious pairing (figure \ref{fig:Vid8Greedy0.3}). 


\begin{figure*}
  \label{fig:Vid19}
  \centering
  \subfloat[Greedy, $\theta = 0.5$.]{
	\includegraphics[height=\linewidth]{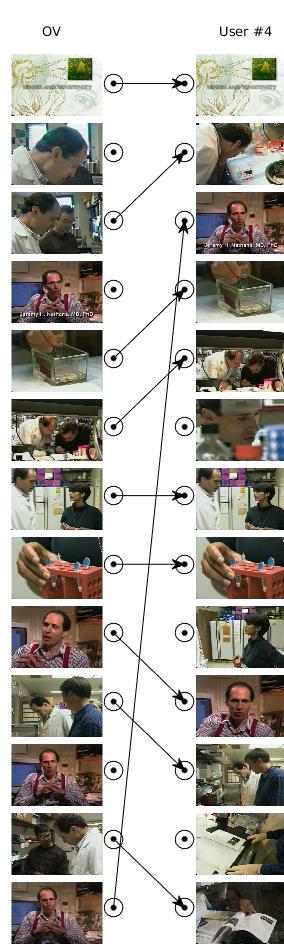}
	\label{fig:Vid19Greedy0.5}
  }
  \subfloat[Kannappan, $\theta = 0.5$.]{
	\includegraphics[height=\linewidth]{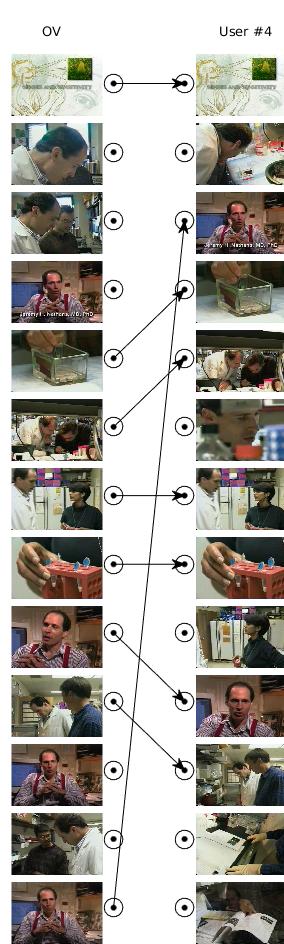}
	\label{fig:Vid19Kannappan0.5}   
  }
  \subfloat[Maximal, $\theta = 0.5$.]{
	\includegraphics[height=\linewidth]{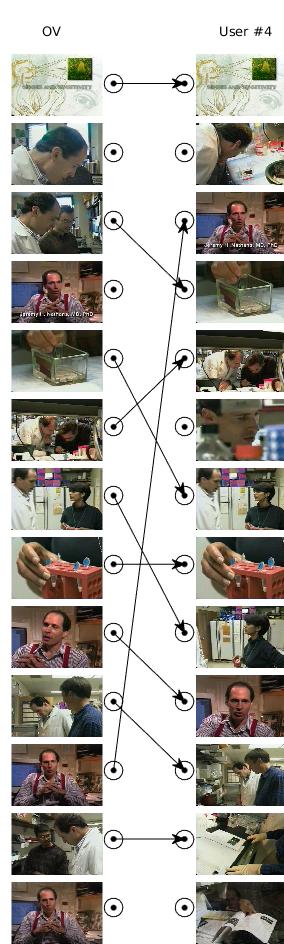}
	\label{fig:Vid19Maximal}
  }
  \caption{Matchings found for Video 29}
\end{figure*}

\begin{figure*}
  \centering
  \subfloat[Greedy, $\theta = 0.5$]{
	\includegraphics[height=0.9\linewidth]{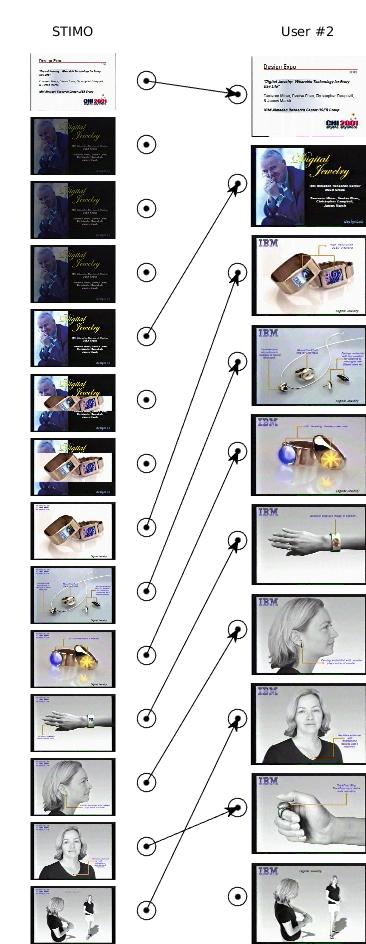}
	\label{fig:Vid14Greedy0.5}
  }
  \subfloat[Kannappan, $\theta = 0.5$]{
	\includegraphics[height=0.9\linewidth]{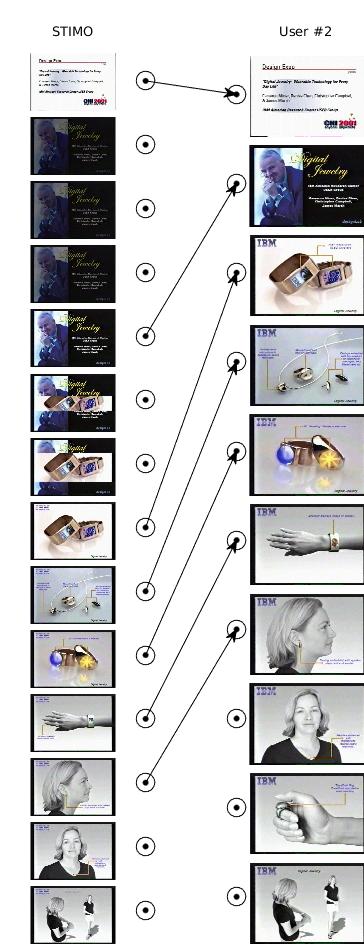}
	\label{fig:Vid14Kannappan0.5}   
  }
  \subfloat[Mahmoud, $\theta = 0.5$]{
	\includegraphics[height=0.9\linewidth]{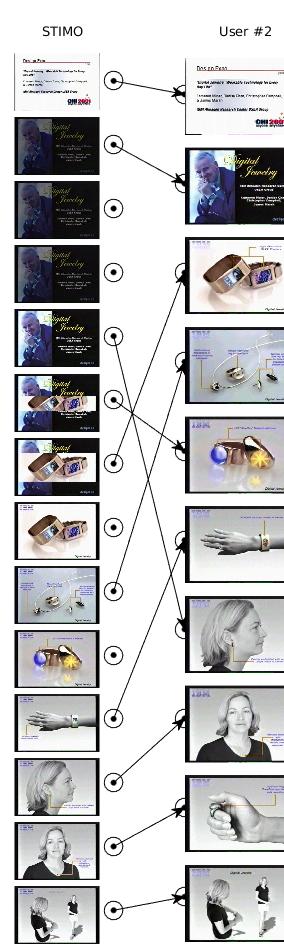}
	\label{fig:Vid14Mahmoud0.5}   
  }
  \caption{Matchings found for Video 34}
  \label{fig:Vid14}
\end{figure*}

Maximal matching algorithms and the algorithm of Mahmoud typically perform much worse than the greedy algorithm or the algorithm of Kannappan et al.  This is true for the first example (video 28, comparing output of OV algorithm to selection of user 2), but we have omitted these matchings for reasons of space.  Example outputs of Mahmoud's algorithm and of a maximal matching algorithm may be seen in figures \ref{fig:Vid19} and \ref{fig:Vid14}.


The example shown in figure \ref{fig:Vid19} illustrates the effect of the fact that the frame ordering is not used in the algorithms (other than Mahmoud's).  The greedy algorithm and the algorithm of Kannappan et al.\ both make a pairing between the final frame of the OV output and a much earlier frame in the selection of user number 4.  This pairing is vastly out of chronological order, but semantically correct: the pairing is between two instances of a recurring concept.  Note that the greedy algorithm again finds one spurious pairing, in addition to the correct pairings found by the algorithm of Kannappan et al., and that the algorithm seeking a maximal matching makes use of some very poor pairings.

Recall that the feature space in all these cases is a Hue histogram: one might expect these matching algorithms to become confused when there is little variation in hue information from one keyframe to the next.  But here again, the conservative approach of the Kannappan algorithm can succeed in avoiding spurious pairings: see figure \ref{fig:Vid14}.

\section{Possible future direction}
\label{sec:directions}
The keyframe matching problem can be viewed as a multi-objective optimization task: to find a matching of greatest possible cardinality, while minimising some function of the weights of the edges.  The thresholding method is only one possible function for penalising high weights, and perhaps it is unnecessarily crude. As it is not \textit{a priori} clear what function of the weights should be used, we propose that sets might be compared by comparing curves of best matchings in a weight-cardinality plane, similar to Receiver Operating Characteristic curves. Hsieh et al.\ \cite{Hsieh95} and Lamb \cite{Lamb98} give algorithms for solving an extension of the bipartite graph matching problem which minimises the sum of the weight of the matching and a penalty for unmatched vertices.  Given a keyframe set for evaluation and a ground-truth, one could repeatedly apply one of these algorithms, incrementally increasing a uniform missed-vertex penalty on each application.  This process would yield the curve we described: the lowest-weight matchings for each cardinality.  The process could perhaps be improved by developing a bespoke algorithm which would not require solving the matching problem from scratch for each new penalty.

\section{Conclusion}
\label{sec:conclusion}
The thresholding procedure developed by de Avila et al.\ allows for use of the information of the distances between frames in a sensible way.  Edges between frames too distant to be matched are removed, which is desirable because if a frame has ``missed'' the other set, it is of no importance just how much it has missed by.  The remaining distance information is retained, which is also desirable - the permissible pairings are not all equal, and it is undesirable to use inferior pairings to increase the number of pairs in the matching.

A simple greedy algorithm is a reasonable way to make use of the remaining distance information following thresholding, but it leaves room for improvement.  The algorithm of Kannappan et al.\ takes to its logical conclusion the idea of seeking the true pairings, not just permissible ones, and seems to achieve better matchings in practice.  In particular, it achieves good results at relatively high values of the threshold cut-off, where the greedy algorithm tends to find spurious pairings.  The conservative approach of the Kannappan algorithm allows the net to be cast wider for possible pairings, without paying a heavy price in the acceptance of poor pairings.

The evaluation method of de Avila et al.\ is entirely sound in concept and we endorse its continued use.  But we recommend the use of the algorithm of Kannappan et al.\ in place of 
the simple greedy algorithm for generating the matching between the keyframe sets.


%



\section*{Acknowledgment}
This work was done under project RPG-2015-188 funded by The Leverhulme Trust, UK.


\end{document}